\pdfoutput=1
\documentclass[11pt]{article}

\usepackage{acl}

\usepackage{times}
\usepackage{latexsym}

\usepackage[T1]{fontenc}
\usepackage[utf8]{inputenc}

\usepackage{microtype}
\usepackage[ruled]{algorithm2e}
\usepackage{microtype}
\usepackage{booktabs}
\usepackage{hyperref}
\usepackage{url}
\usepackage{amsmath}
\usepackage{graphicx}
\usepackage{subcaption}
\usepackage{multirow}
\usepackage{inconsolata}
\usepackage{float}
\usepackage{subcaption}

\title{Stateful Memory-Augmented Transformers for Efficient Dialogue Modeling}

\author{Qingyang Wu \\ Columbia University \\        \texttt{qw2345@columbia.edu}
        \And Zhou Yu \\ Columbia University \\  \texttt{zy2461@columbia.edu}}
\begin{document}
\maketitle

% TODO: add more discussion about Memformer

% Edited times: 5
\begin{abstract}

% Existing Problems
Transformer encoder-decoder models have achieved great performance in dialogue generation tasks, however, their inability to process long dialogue history often leads to truncation of the context
% What we do -> add recurrence
To address this problem, we propose a novel memory-augmented transformer that is compatible with existing pre-trained encoder-decoder models and enables efficient preservation of the dialogue history information.
% what does it do
By incorporating a separate memory module alongside the pre-trained transformer, the model can effectively interchange information between the memory states and the current input context.
%% Dual attention stream
% It incorporates a dual attention stream to effectively interchange information between the memory module and the pre-trained Transformer and a gating mechanism to retain history information. 
% Results
We evaluate our model on three dialogue datasets and two language modeling datasets. 
Experimental results show that our method has achieved superior efficiency and performance compared to other pre-trained Transformer baselines.

\end{abstract}

% Edited times: 0
\section{Introduction}

% TODO: make stronger highlights for latency.

\begin{figure}[ht]
    \centering
    \begin{subfigure}[b]{0.46\textwidth}
    \centering
        \includegraphics[width=0.8\columnwidth]{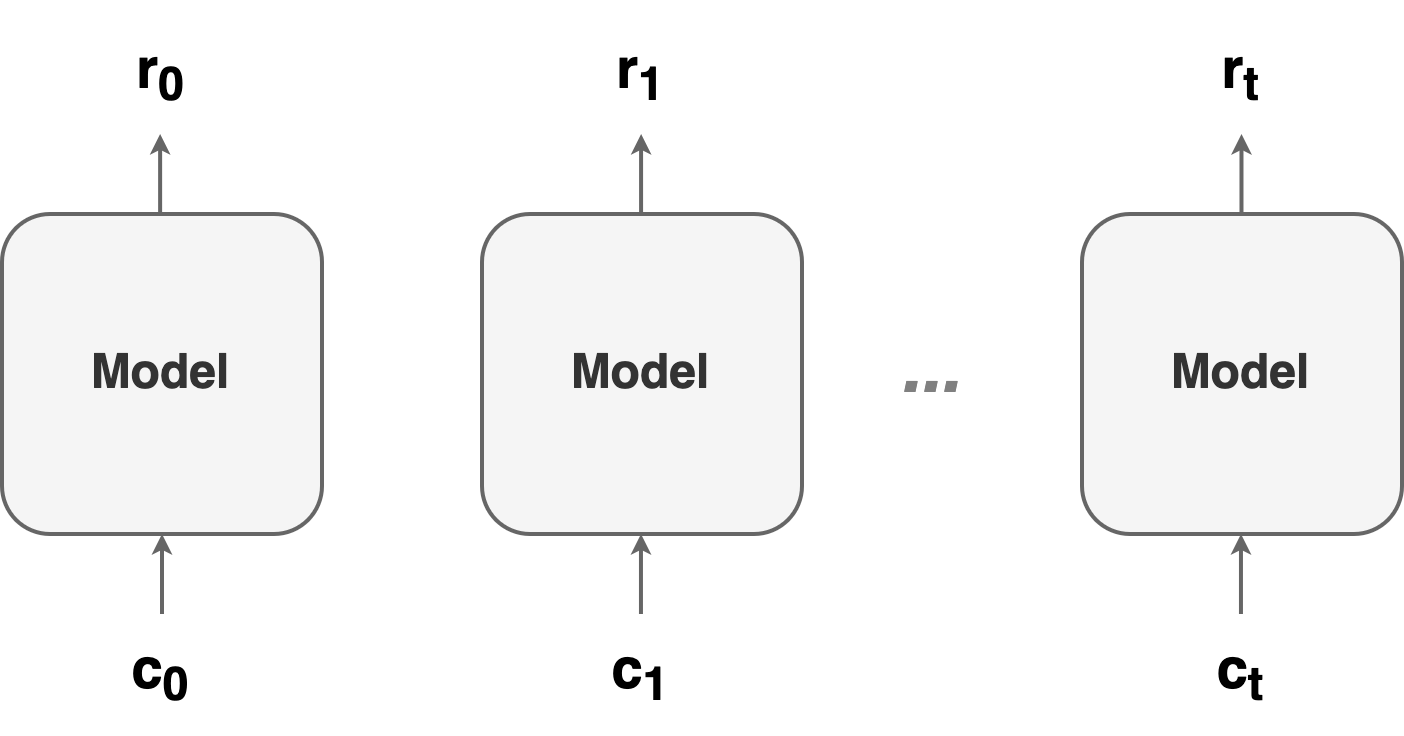}%
        \caption{Stateless model: history information can only be inferred from context. }
    \end{subfigure}
    
    \begin{subfigure}[b]{0.46\textwidth}
        \centering
        \includegraphics[width=0.8\columnwidth]{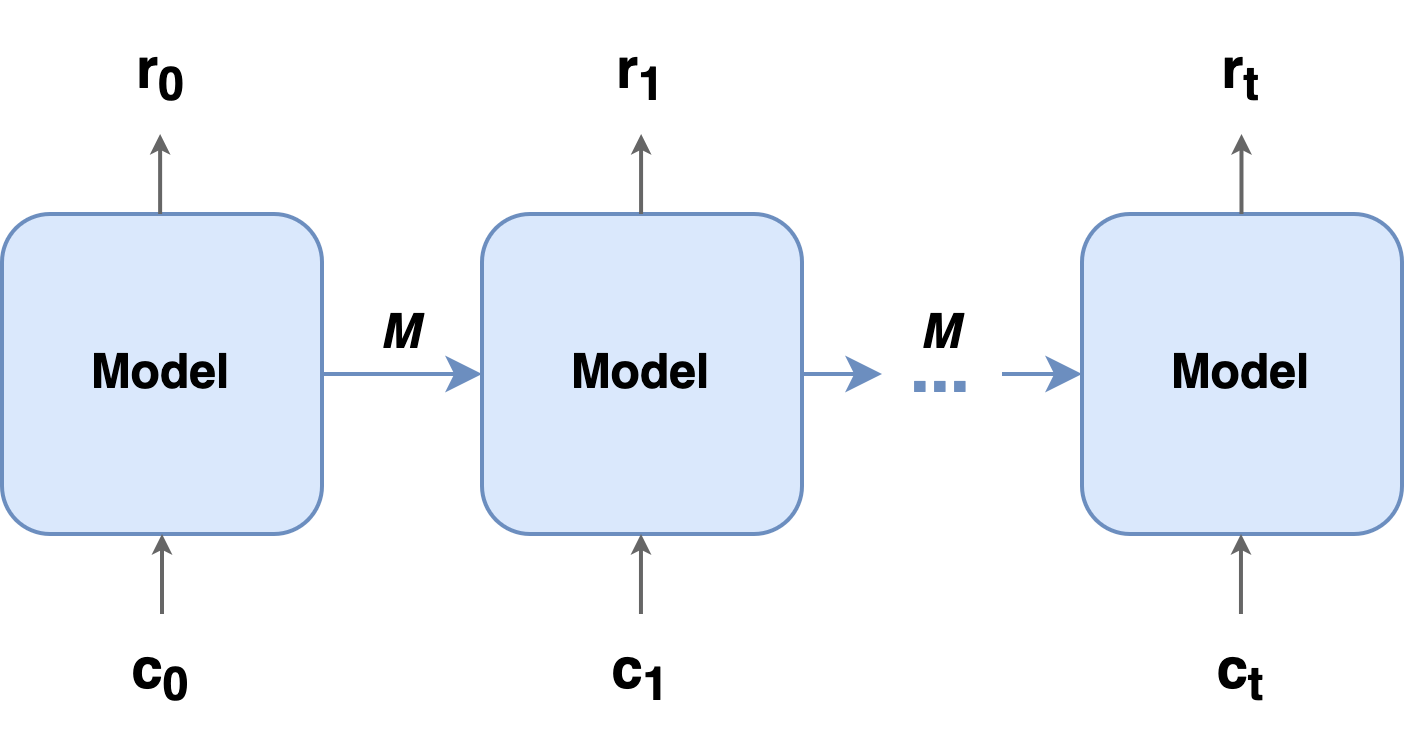}%
        \caption{Stateful model: history information is carried by memory states $M$.}
    \end{subfigure}
    \caption{Illustration of Stateful vs. Stateless. ``State" means a model's internal state representations. $c_t$ and $r_t$ represent the dialog context and response at timestep $t$. Stateful models can have smaller context size compared to stateless models because of memory.
    % Most existing Transformer encoder-decoder models are Stateless.
    }
    \label{fig:stateful}
\end{figure}

% Background of Transformers encoder-coder and  Dialog Modeling
% Edited times: 2
Recently, Transformers \cite{DBLP:conf/nips/VaswaniSPUJGKP17} have achieved state-of-the-art results in many natural language processing tasks, particularly in language understanding and generation. 
In the field of open-domain dialogue modeling, DialoGPT \cite{DBLP:conf/acl/ZhangSGCBGGLD20} has achieved great performance by extending the Transformer decoder model GPT2 \cite{radford2019language} by pre-training it on a large corpus of open-domain dialogues. 
Subsequently, Meena \cite{DBLP:journals/corr/abs-2001-09977} and BlenderBot \cite{DBLP:conf/eacl/RollerDGJWLXOSB21} further improved the performance of response generation with larger Transformer encoder-decoder models.

% Weaknesses & Problem
% Edited times: 2
However, the attention mechanism in Transformer-based dialogue models, which has complexity scaling quadratically with the sequence length, makes them computationally expensive for long context inputs. 
As an example, BlenderBot \cite{DBLP:conf/eacl/RollerDGJWLXOSB21} has to truncate the input length to 128 tokens for better efficiency, otherwise, the model's computational cost would become infeasible for real-time conversation tasks such as chatbot applications.

% Related work to address the problem
% Edited times: 2
Many studies have addressed the challenge of processing long sequences with Transformers \cite{DBLP:conf/icml/KatharopoulosV020,DBLP:journals/corr/abs-2202-08791,DBLP:journals/corr/abs-2202-10447,DBLP:conf/acl/DaiYYCLS19,DBLP:conf/iclr/RaePJHL20}.
% Problem 1
However, they focused on pure language modeling tasks and are primarily decoder-only models.
% Problem 2
Another limitation is that their models are not pre-trained with large corpora, which increases difficulty for performance comparison with existing pre-trained Transformers.
% Longformer
More recently, \citet{DBLP:journals/corr/abs-2004-05150} addressed the problem by proposing Longformer Encoder-Decoder (LED) based on the pre-trained encoder-decoder model BART \cite{DBLP:conf/acl/LewisLGGMLSZ20} for sequence-to-sequence tasks.
% Longformer's problem
It uses a sparse attention window and achieves a linear time complexity.
Nevertheless, LED is inefficient in dialogue modeling, because it is stateless and depends on the context to provide history information.
% while the context needs to be re-processed by the encoder for every utterance.

% Stateful vs Stateless & Figure 1 explanation
% Edited times: 3
In this work, we utilize the idea of Memory-Augmented Transformer (Memformer) \cite{DBLP:journals/corr/abs-2010-06891} and convert an existing pre-trained Transformer into a stateful model with internal memory representations.
% Motivation
A stateful model can keep history information in its internal hidden states in contrast to a stateless model.
% Stateless Model
As shown in Figure~\ref{fig:stateful}, most existing Transformer encoder-decoder models are stateless. 
They rely on the input context to provide history information, and therefore they typically require a larger context to avoid information loss.
% Stateful Model
For a stateful model, it can store history information in its memory states. 
With a smaller context size, the stateful model can still retain most of the history information, which results in better efficiency than a stateless model.

% How Memformer maintains the state
% Edited times: 2
Memformer \cite{DBLP:journals/corr/abs-2010-06891} achieves statefulness by having internal memory states to store history information.
The memory size is fixed so that the model will prioritize memorizing important information.
% What is inside
To interact with the memory, it consists of a memory reader and a memory writer into a Transformer encoder-decoder model.
% Result
Memformer has shown better efficiency on the language modeling dataset WikiText-103 \cite{DBLP:conf/iclr/MerityX0S17} than the decoder-only models Transformer-XL \cite{DBLP:conf/acl/DaiYYCLS19} and Compressive Transformer \cite{DBLP:conf/iclr/RaePJHL20}.
% Memformer weaknesses
However, Memformer only focused on language modeling tasks and was not pre-trained on large corpora, and hence it cannot be used for downstream applications.
Also, its structure does not fit the existing pre-trained Transformer encoder-decoder models.

% What we do to improve Memformer
% Edited times: 4
To address these limitations of Memformer, we propose MemBART with new architecture modifications and training techniques that can convert the existing pre-trained Transformer encoder-decoder model BART \cite{DBLP:conf/acl/LewisLGGMLSZ20} into a stateful memory-augmented Transformer encoder-decoder model.
% Dual Attention Stream
Specifically, we introduce a dual attention stream to enhance the memory module, which is accomplished by using a separate Transformer to update the memory states at each layer.
% Gating Mechanism
We also implement a residual gated memory update mechanism to better retain important history information.
At each timestep, the gating mechanism controls the extent of keeping or overwriting each memory slot's values for the next timestep.
% Pre-training
We further pre-train the memory module and enable the model to memorize important history information.
As MemBART is a pre-trained model, it can be used for broader downstream applications.

% Conclusion Part
% Edited times: 2
% Contribution about our approach
Our contributions focus on introducing a novel stateful memory-augmented Transformer encoder-decoder model that is compatible with the existing pre-trained language model BART.
% Evaluation
We evaluate our model's effectiveness on three dialogue datasets and two language modeling datasets.
% Result
Experimental results demonstrate our model's superior efficiency in terms of latency and performance.
% Some additional part
We will release the checkpoints of our pre-trained MemBART models.

% Edited times: 3
\section{Related Work}

% One direction is replace softmax in attention with linear kernel. 
% Linear Transformer \cite{DBLP:conf/icml/KatharopoulosV020}, CosFormer \cite{DBLP:journals/corr/abs-2202-08791}, FLASH \cite{DBLP:journals/corr/abs-2202-10447}. 

% Transformer XL \cite{DBLP:conf/acl/DaiYYCLS19}, Compressive Transformer.

\subsection{Stateful Neural Networks}

% RNNs are Stateful
Recurrent neural networks (RNN) are naturally stateful models.
% Explanation
Training RNNs on long time-series data often requires truncated back-propagation through time \cite{DBLP:journals/neco/WilliamsP90} and passing the internal states of the model to the next batch.
Stateful RNNs are also widely used for recurrent reinforcement learning \cite{DBLP:conf/cifer/Gold03,DBLP:conf/aaaifs/HausknechtS15}, where the states of the agent need to be maintained.
There have been variants of stateful RNNs \cite{DBLP:journals/corr/WestonCB14,DBLP:conf/nips/SukhbaatarSWF15,DBLP:journals/nature/GravesWRHDGCGRA16} studied to solve various tasks.
However, due to parallel inefficiency, they are gradually succeeded by large Transformer models \cite{DBLP:conf/nips/VaswaniSPUJGKP17}.

% Decoder-only
Decoder-only Transformers can be stateful by storing the previously computed keys and values.
Transformer-XL \cite{DBLP:conf/acl/DaiYYCLS19} and Compressive Transformer \cite{DBLP:conf/iclr/RaePJHL20} explore this direction, but their states have a theoretical maximum range of maintaining the information from previous tokens.
Thus, they normally require a large memory size to be effective.

Linear attention Transformers can act as RNNs with states. 
They use a linearized kernel to approximate softmax operation.
Different variants of linear Transformers \cite{DBLP:conf/icml/KatharopoulosV020,DBLP:journals/corr/abs-2202-10447,DBLP:journals/corr/abs-2202-08791} have been proposed and achieved great performance in language modeling tasks.
% However, linear attention loses generality over varied sequence lengths without normalization using softmax.
However, there are no pre-trained large linear Transformers yet.
Similar models such as Memorizing Transformer \cite{wu2022memorizing}, Block-Recurrent Transformer \cite{DBLP:journals/corr/abs-2203-07852} all focus only on language modeling tasks and are not applicable for other downstream tasks.

% \subsection{Encoder-Decoder Models}
% LongT5. Longformer LED
% As longfomer is also based on BART. We use LED as a strong baseline.

\subsection{Stateless Long-Document Models}

% Sparse Transformer
For long documents processing, sparse Transformers are another direction.
The main idea is to apply a sparse attention matrix to skip computations of tokens that are far away.
Many works \cite{DBLP:journals/corr/abs-1904-10509,DBLP:conf/nips/ZaheerGDAAOPRWY20,DBLP:journals/corr/abs-2004-05150} have explored different sparse attention patterns with linear complexity.
% Most of them have linear computation complexity.
Especially, Longformer extended the pre-trained BART \cite{DBLP:conf/acl/LewisLGGMLSZ20} with sparse attention and introduced Longformer-Encoder-Decoder (LED) for sequence-to-sequence tasks.
However, these models are stateless, which are inefficient for dialogue modeling.
They require the context to be long enough to cover enough history information.
The context also needs to be re-computed at every timestep due to bidirectional attention.
Besides, sparse Transformers need full attention for the local window, which makes them less competitive against non-sparse models when the context is short.
In contrast, our stateful memory-augmented method can have a shorter context input while still memorizing the history information.

% \subsection{Encoder-Decoder Models}

% Encoder-decoder models such as BART \cite{DBLP:conf/acl/LewisLGGMLSZ20} and T5 \cite{DBLP:journals/jmlr/RaffelSRLNMZLL20} have demonstrated better performance compared to decoder-only models such as GPT2 trained with causal language modeling. More recently, an improved encoder-decoder UL2 with 20B parameters has outperformed GPT3 with 175B parameters in both fine-tuning and few-shot settings.

% % Few works studied how to 
% Few works studied how to build encoder-decoder Transformers for long squences, which can benefit applications such as dialogues, summarization, or abstractive question answering.
% % Longformer enhances the pre-trained BART encoder-decoder model.
% In this paper, we explore the direction of applying Memformer techniques to existing large pre-trained Transformer encoder-decoder models.

\begin{figure*}[t]
    \centering
    \includegraphics[width=0.9\textwidth]{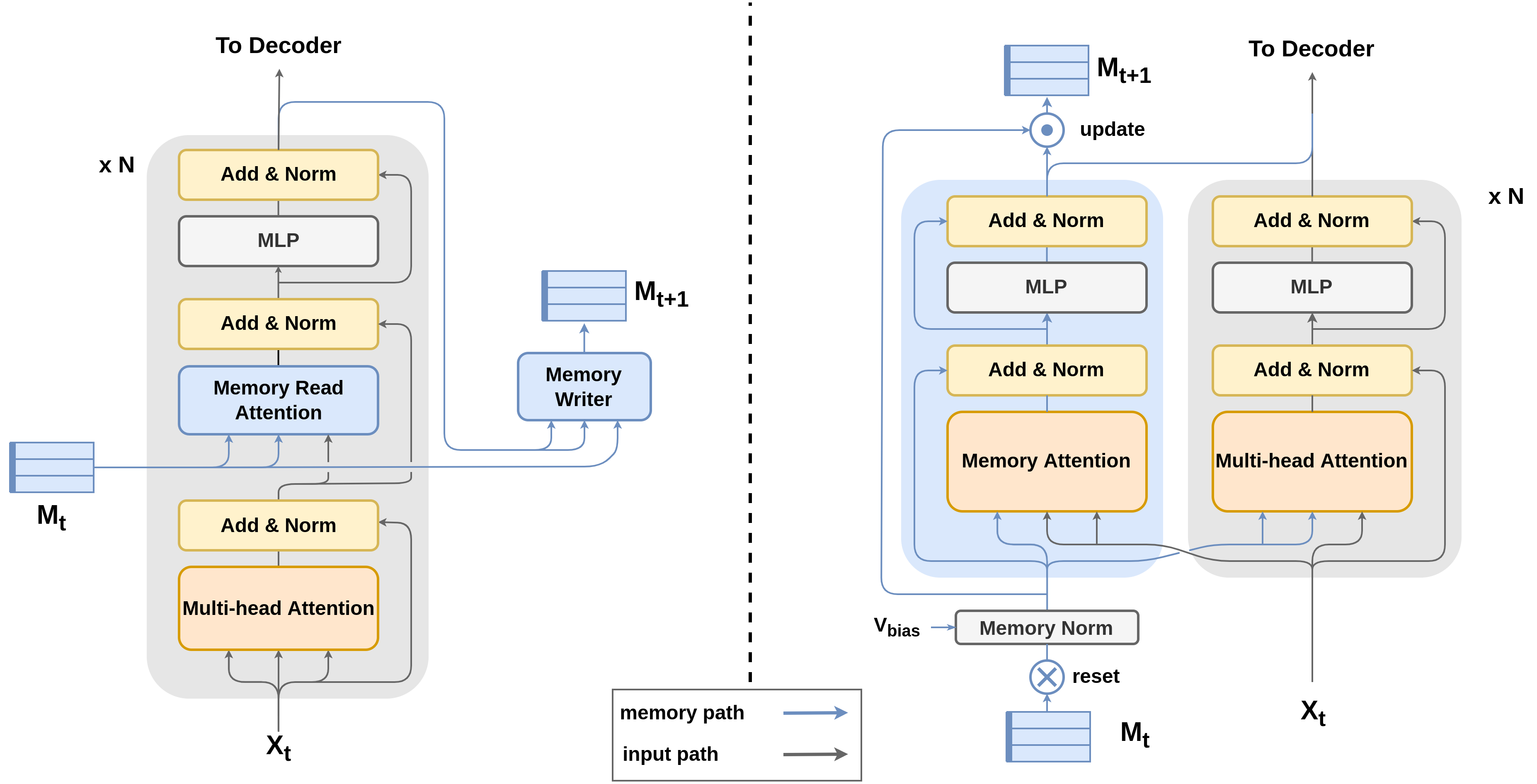}
    \caption{\textbf{Left}: Memformer with cross attention to read from memory and a separate memory writer to update information in memory slots.  \textbf{Right}: MemBART with the dual attention stream to handle memory reading and writing simultaneously. This design reduces the interference with the pre-trained model's distribution.}
    \label{fig:architecture}
\end{figure*}

\section{Methods}

% What is done first
In this section, we first describe the background of memory-augmented Transformers.
% Then
Then we introduce an novel memory module that is compatible with existing Transformer encoder-decoder models. 
% Pre-training
We further pre-train the memory module with the sequence denoising objective to initialize the memorization capability.
% Analysis
In the end, we analyze the theoretical complexity of our proposed model for dialogues.
% % Data Dispatcher
% In the end, we cover the different batch processing in our stateful paradigm.

\subsection{Memory-Augmented Transformer}

% What Memformer does
Memformer \cite{DBLP:journals/corr/abs-2010-06891} modifies a Transformer encoder to interact with a fixed-size dynamic memory, so that it can store and retrieve history information.
It comprises a memory reader and a memory writer.
% In detail
The memory reader utilizes cross attention to retrieve history information from the memory $M_t$:
\begin{align*}
    Q_{H^l}, K_{M^l}, V_{M^l} & = H^l W_Q, M_t W_K, M_t W_V         \\
    A^{l}     & = \text{MHAttn} (Q_{H^l}, K_M)        \\
    H^{l+1}           & = \text{Softmax}(A^{l}) \, V_M
\end{align*}
where $H^l$ is the input's hidden states at layer $l$.

For the memory writer, each memory slot $m^i_t \in M_t$ is projected into a query to attend to itself and the final layer's input hidden states $H^L$:
% This process determines how much old information to keep in the memory slot versus the new information from the input sequence.
\begin{align*}
    Q_{m^i_t},  K_{m^i_t} &= m^i_t W_Q, m^i_t W_K                         \\
              K_{H^L}, V_{H^L} &= H^L W_K, H^L W_V                            \\
             A_{m^i_t} =&  \text{MHAttn} (Q_{m^i_t}, [K_{m^i_t}; K_{H^L}]) \\
             m^i_{t+1} =& \text{Softmax}(A^{m^i_t}) [m^i_t ;  V_{H^L} ]
\end{align*}

Memory states are reset with the reset signal $r$. 
\begin{align*}
     r &= 
    \begin{cases}
    1, & \text{if } t=0 \\
    0              & \text{otherwise}
    \end{cases} \\
     M'_t &= \text{LayerNorm} ( (1 - r) \odot M_t + v_b) 
    %  M'_0 &= \text{LayerNorm} (v_b)
\end{align*}
% Memory Initialization with bias
Also, we normalize the memory states at every timestep with a bias term $v_b$ as the forgetting mechanism. $v_b$ determines the initial memory $M_0$ which is $\text{LayerNorm} (v_b)$.

\subsection{Dual Attention Stream}

%Previous problem
Memformer adds cross-attention layers between self-attention and feed-forward layers to achieve memory functionality.
% Problem
However, directly injecting layers inside a pre-trained Transformer will interfere the distribution of learnt knowledge and lead to worse performance.
% Motivation & Challenge
Therefore, we aim to integrate the memory module with a minimal influence of the original pre-trained Transformers.
% Thus, we need to replace the memory reading and writing modules so that they will not influence the original distribution.

% Our approach
We propose a dual attention stream so that the memory path has minimal interference with the input sequence's data path.
% What we do in detail
Inside every layer $l$, we separately project the input sequence $H^l$ and the memory states $M^l$ to  queries $Q$, keys $K$, and values $V$:
\begin{align*}
    Q_{H^l}, K_{H^l}, V_{H^l} &= W_{H^l} H^l \\
    Q_{M^l}, K_{M^l}, V_{M^l} &= W_{M^l} M^l
\end{align*}

Then, there are two attention streams to realize memory reading and memory writing simultaneously at each layer:
\begin{align*}
    A_{H^l} &= \text{Attention} (Q_{H^l}, [K_{M^l};K_{H^l}]) \\
    H^{l+1} &= \text{Softmax} (A_{H^l}) [V_{M^l}; V_{H^l}] \\
    A_{M^l} &= \text{Attention} (Q_{M^l}, [K_{M^l};K_{H^l}]) \\
    M^{l+1} &= \text{Softmax} (A_{M^l}) [V_{M^l}; V_{H^l}]
\end{align*}

% Memory reading stream
Specifically, the attention stream $A_{H^l}$ serves as memory reading, where the input sequence's hidden states $H^l$ gathers the information from the memory states $M_t$ to get the next layer's representation $H^{l+1}$.
% Memory writing stream
The other attention stream $A_{M^l}$ serves as memory writing. 
Note that we update memory states at every layer.
Each memory slot $m^l \in M^l$ attend to itself and the input's hidden states to obtain the next layer's memory slots $M^{l+1}$.
Each memory slot does not interfere with other memory slots when updating.

% Advantage
This dual attention stream allows the information to exchange effectively between the memory slots and the input sequence, while minimally affects the original pre-trained Transformer's knowledge.

% The overall architecture and difference is shown in Figure~\ref{fig:architecture}.

\subsection{Residual Gated Memory Update}
% Problem & Motivation
The dual attention stream achieves memory reading and writing simultaneously at each layer.
However, as the number of layers increases, the final layer's memory representation may be hard to retain the previous timestep's information.

As a workaround, we implement a residual gating mechanism. 
% similar to gated recurrent unit \cite{DBLP:journals/corr/ChungGCB14}.
We let the encoder predict a score $z_t \in (0, 1)$ with sigmoid to control the update of each memory slot separately.
% It controls how much information to keep or discard at each time step.
\begin{align*}
    H_{M_{{t+1}}} &= \text{Encoder} (x_t, M_t) \\
    M'_{t+1} &= \text{MLP} (H_{M_{{t+1}}}) \\
    z_t &= \sigma_z ( W_z H_{M_{{t+1}}} + b_z) \\
    M_{t+1} &= z_t \odot M'_{t+1} + (1 - z_t) \odot M_t
\end{align*}

$x_t$ is the input sequence length. 
$H_{M_{{t+1}}}$ is the final layer's memory hidden states.
$M'_{t+1}$ is the next timestep's memory slots candidate.
% $z_t$ is controls the update of each memory slot separately.

% Edited times: 1
\subsection{Learning to Memorize Important Information}

% Motivation
As the memory size is fixed, the model needs to learn what information to keep and what to forget, but the memory module initially has no knowledge of that.
Therefore, it requires further pre-training for the memory module to learn to memorize important information.

% What we do
We use the sequence denoising objective as the memory module's pre-training objective.
% Understand the benefit.
We split a document into segments, add random masks to these segments, and feed them into the model sequentially.
% Why it works
This objective can teach the model to memorize important information.
If important words such as named entities appear in previous timesteps but are masked in the current input context, the model can predict them back with the help of memory.
For less important words that can be easily inferred from the context or grammar, the model can choose not to store them in the dynamic memory.

% % Text recall Motivation
% TODO: make it more fluent
% In addition, we introduce a task called text recall.
% It is used to test the memory module's effectiveness.
% We also use it in final pre-training, with a probability ($10\%$), the model needs to predict the previous timestep's text input.
% The task helps the model to better retain the history information.

% Edited times: 3
\subsection{Complexity Analysis}

% TODO: Consider Longformer type of models where their complexity is O(T^2 N)

% Memformer overall advantage
Our method is efficient in processing long sequences compared to traditional Transformers, especially in modeling dialogues.
% Analysis
For example, consider a dialogue with $T$ turns, and $N$ tokens at each turn.
% Decoder-Only
The overall complexity for a Transformer to process all the turns would be $\mathcal{O}( N^2 + 2N^2 + \ldots + TN^2)$, or simply $\mathcal{O}(T^2N^2)$. 
% Encoder-decoder
If we keep all the history tokens, a traditional encoder-decoder model would require to re-compute all the history tokens because of the bidirectional attention, which increases the complexity. 
% For uni-directional models that can store the previously computed activations, the complexity is still $\mathcal{O}(T^2N^2)$ due to the attention mechanism. 
In practice, due to the limitation of the maximum number of positional embeddings and the GPU memory constraint, we often truncate the dialog history to a fixed length.

% but it still holds that the encoder-decoder and decoder-only models will have a quadratic complexity with respect to the number of turns $T$.

% Our method
In contrast, our stateful model can store the history information in the fixed-size memory.
The implementation has a complexity of $\mathcal{O}(T N^2)$, and it does not require re-computation for the history tokens.
% Longformer
For efficient Transformer models such as Longformer, the complexity can be reduced from $\mathcal{O}(T^2N^2)$ to $\mathcal{O}(T^2 N)$. 
However, when the context length $N$ is small, the number of turns $T$ is the leading factor for efficiency, where our method shows better efficiency in theory.

% With sparse memory attention pattern (each memory slot only attends to itself), it further optimizes the cost to $\mathcal{O}(T (MN + N^2))$, which is more efficient when the memory size is large.

\section{Memory Module Pre-training}

% Motivation
As mentioned above, the memory module needs to be pre-trained to learn to memorize important information.
However, to compare the effectiveness of our proposed approach with the previous models, it would be expensive to pre-train all model variants.
% What we do
Therefore, we use a simple text recall task to evaluate different models before pre-training on large corpora.

% How we initialize the model
For all model variants, we choose BART \cite{DBLP:conf/acl/LewisLGGMLSZ20} as the backbone as it has demonstrated great performance on conversational datasets. 
We also initialize the memory module's self attention and feed-forward parameters with the pre-trained weights for better adaptation.

% However, without further pre-training, the memory module does not learn to focus on capturing important information in the fixed-size memory.
% Simple copy task
\subsection{Model Selection with Text Recall Task}

\begin{figure}[h]
    \centering
    \includegraphics[width=0.9\columnwidth]{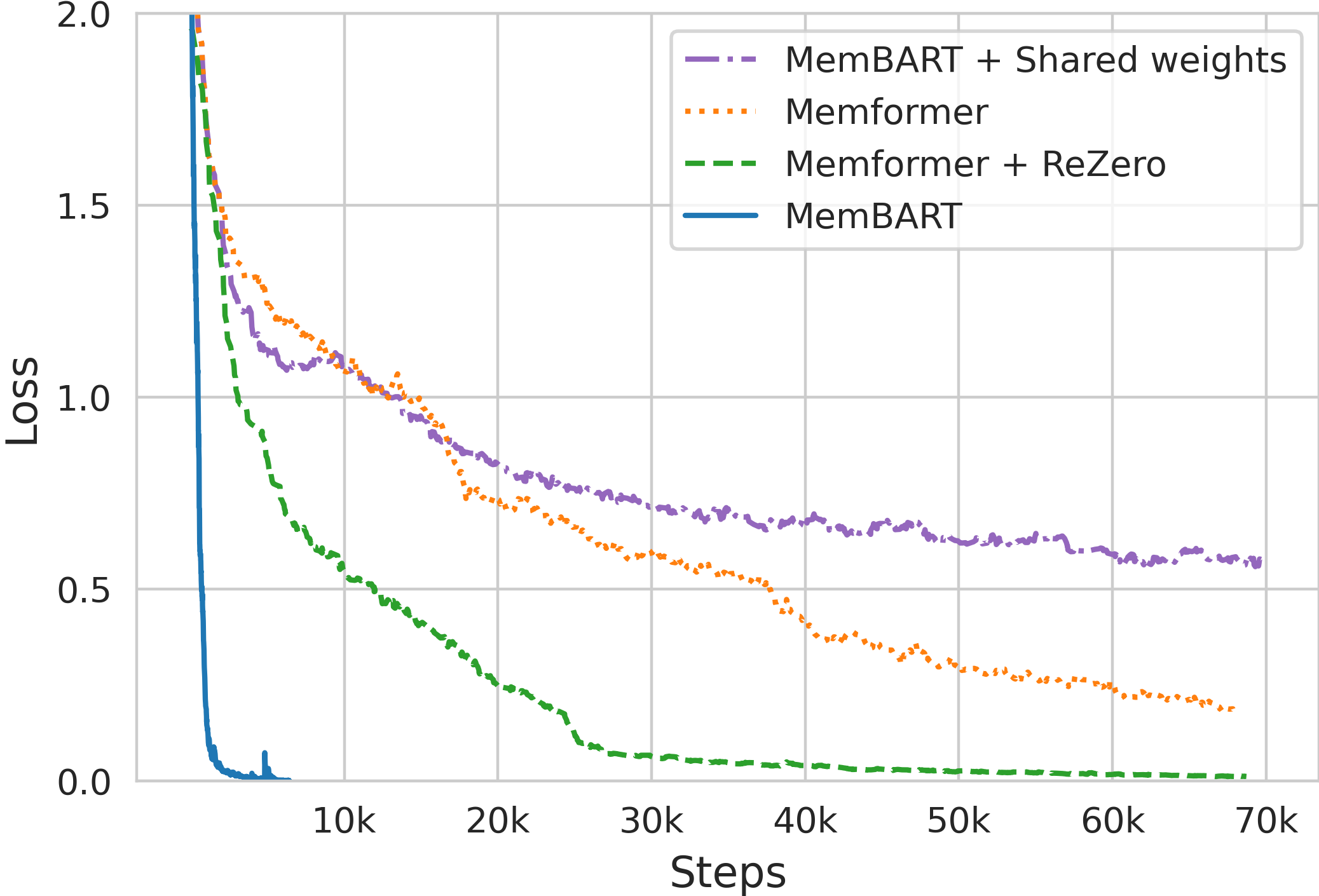}
    \caption{Loss curves for different models for the text recall task.}
    \label{fig:text_recall_loss}
\end{figure}

% Task description
The text recall task lets the model recover the previous timestep's input text, where the history information can only flow through the memory bottleneck.

We evaluate different model variants with the text recall task to select the best model before pre-training. 
The first is directly adding the memory cross-attention layers into BART (Memformer), which the model's architecture is similar to Memformer \cite{DBLP:journals/corr/abs-2010-06891}.
The second model uses ReZero \cite{DBLP:conf/uai/BachlechnerMMCM21} that it applies a zero-initialized trainable weight when adding the memory cross-attention layer, so that the model's output distribution is not changed initially (Memformer + ReZero).
The third model is our proposed MemBART where the memory module shares the weights with BART (MemBART + Shared weights).
The last one is our final model MemBART without sharing weights between the memory module and the pre-trained Transformer (MemBART).

% Results
The training details are in Appendix~\ref{appx:text_recall_pretrain}.
In Figure~\ref{fig:text_recall_loss}, we can observe that the original Memformer (orange) did not converge to zero loss. 
MemBART with shared weights (purple) also did not converge and performed worse, suggesting that the memory states should have different distribution space from the word embeddings.
Memformer with ReZero (green) converged slowly in the end.
In comparison, MemBART (blue) only used one quarter of the time to reach nearly zero loss.
% Result
The result shows that our proposed memory module architecture is compatible with the pre-trained BART and can be efficiently trained for memorization tasks.

\begin{table*}[!htb]
    \centering
    \resizebox{0.86\textwidth}{!}{
        \begin{tabular}{l|cc|cc|cc|cc}
        \toprule
\multirow{2}{*}{\textbf{Models \quad \textbackslash \quad Context}} & \multicolumn{2}{c|}{\textbf{64}} & \multicolumn{2}{c|}{\textbf{128}} & \multicolumn{2}{c|}{\textbf{256}} & \multicolumn{2}{c}{\textbf{512}} \\
& \textbf{PPL} $\downarrow$ & \textbf{F1} $\uparrow$  &\textbf{PPL}$\downarrow$  & \textbf{F1} $\uparrow$ & \textbf{PPL}$\downarrow$  & \textbf{F1} $\uparrow$ & \textbf{PPL}$\downarrow$  & \textbf{F1} $\uparrow$ \\
\midrule
BART base & 10.91&	25.01&	9.39&	25.44&	8.64&	26.31&	8.76&	26.22 \\
% BART base* & & -&	-&	-&	-&	-&	-&	- \\
MemBART base (64)* &	8.68 &	27.34&	8.58 &	27.37&	8.46&	27.05&	-&	- \\
\quad w/o history &	10.52 &	25.54 &	9.44 &	26.52 &	8.57 &	26.23 &	-&	- \\
\quad w/o pre-training &	10.67 &	25.26 &	9.37 &	26.12 &	8.60 &	26.45 &	-&	- \\
MemBART base (128) & \textbf{8.59}&	27.45&	8.57 &	27.52&	8.39&	\textbf{27.52}&	-&	- \\
MemBART base (256) & 8.60&	\textbf{27.65} & \textbf{8.49} &	\textbf{27.68}&	\textbf{8.38}&	27.41&	-&	-\\
\bottomrule
\toprule
GPT2-12 & 10.93&	25.18&	9.86&	26.03&	9.06&	26.55&	9.04&	26.52 \\
GPT2-24 & 9.51&	25.46&	8.56&	26.52&	7.82&	27.19&	7.81&	27.20 \\
BART large & 9.12&	25.50&	8.01&	26.84&	7.33&	28.67&	7.31&	28.64  \\
MemBART large (128) & \textbf{7.47} &	\textbf{28.06}&	\textbf{7.33}&	\textbf{28.57}&	\textbf{7.15} &	\textbf{29.16} & - & - \\
        \bottomrule
        \end{tabular}
    }
    \caption{PersonaChat results. We report perplexity (PPL) and F1 with different context lengths.
    * MemBART (64) means the memory size is 64. ``w/o pre-training" means without pre-training the memory module.}
    \label{tab:personachat}
\end{table*}

% Edited Times: 1
\subsection{Sequence Denoising Pre-training}

% Objective
We have shown that the proposed MemBART has outperformed Memformer and other model variants.
Now, we pre-train MemBART with the sequence denoising objective for the memory module to memorize important information.
We have two sizes of models: MemBART base (183M) and MemBART large (558M).
% Corpus
We use a similar pre-training corpus to BART to avoid data leaking, which includes a subset of BooksCorpus \cite{DBLP:conf/iccv/ZhuKZSUTF15}, CommonCrawl \cite{DBLP:journals/jmlr/RaffelSRLNMZLL20}, OpenWebText \cite{Gokaslan2019OpenWeb}.
We filter out documents that are less than $512$ tokens for better memory learning.
% Hyper-parameters
We split the document into segments with a window size of $512$ and an overlap of $128$ tokens.
At each timestep, we randomly mask $30\%$ of input sequence tokens.
% We use memory replay back-propagation for the large model.
We also develop a novel batch processing technique mentioned in Appendix~\ref{appx:batch-processing} to handle the temporal dependency between batches.
Other pre-training details are in Appendix~\ref{appx:pre-training}.

\begin{figure}[h]
    \centering
    \includegraphics[width=0.9\columnwidth]{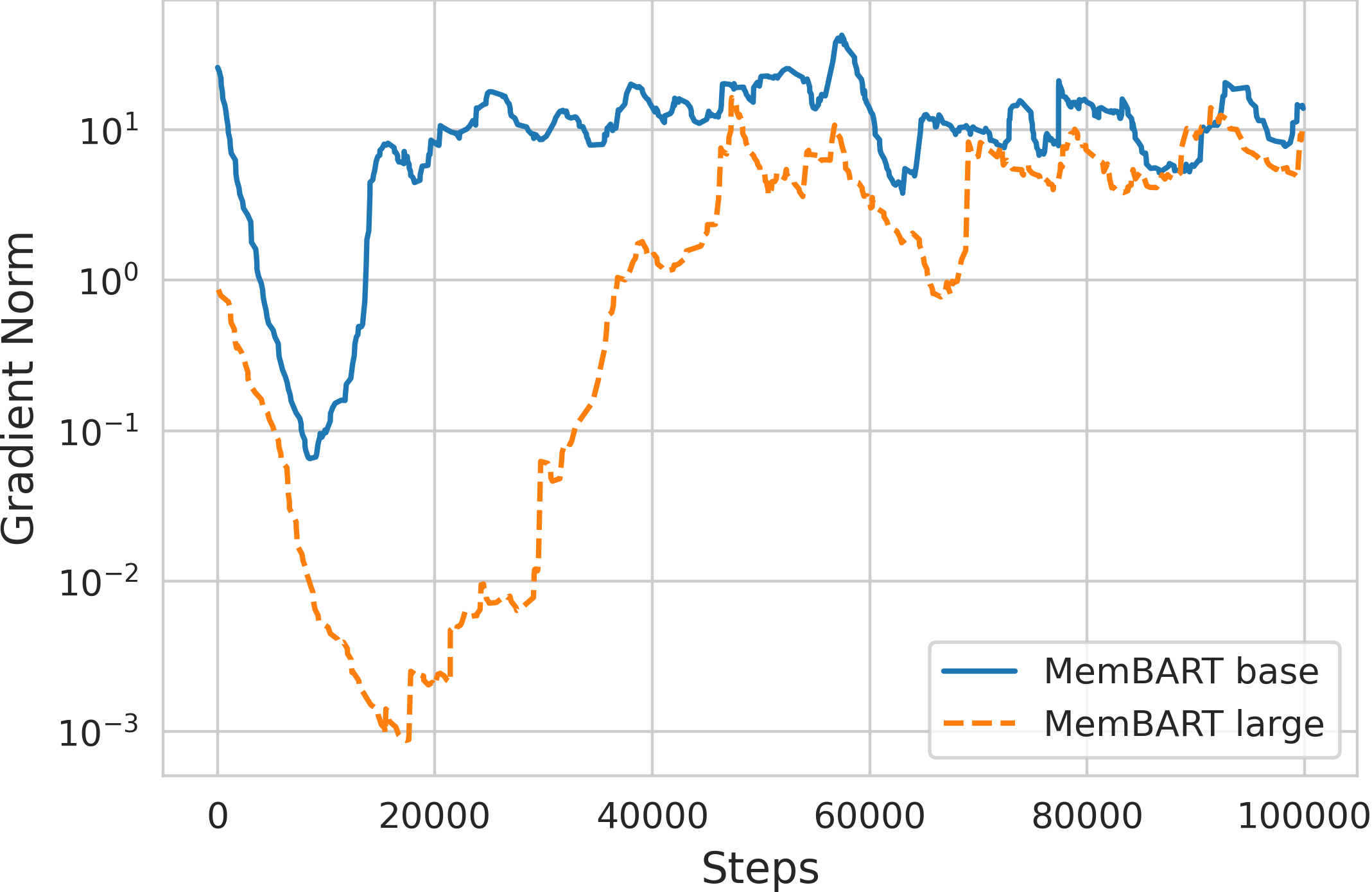}
    \caption{Memory's gradient norm during pre-training. When the gradient is near the minimum, the model performs terribly in downstream tasks.}
    \label{fig:gradient_norm}
\end{figure}

In Figure~\ref{fig:gradient_norm}, we show the magnitude of the gradients flowing through memory states during pre-training.
% MemBART base
At the early stage of the pre-training (less than 20,000 steps), we observe that the MemBART base model does not perform well in the downstream tasks.
We suspect that when the gradient norm is small, it means that model is not actively using the memory states.
Therefore, the gradient norm serves as an indicator of when the memory module is learnt.
% MemBART large
For MemBART large, the downstream tasks' performance improves after 50,000 steps when the gradient norm reaches the maximum.
This pattern suggests that it needs a certain number of pre-training steps for the memory module to learn to memorize important information, and the large model needs more update steps to learn memorization.

% Variance
% We run the experiments with three random seeds.
% The variance for perplexity is smaller than $0.1$. 
% For $F1$, it is smaller than $0.5$.

\section{Downstream Applications}

In this section, we introduce the downstream applications and datasets for evaluation. Then, we show the results on the dialogue and language modeling tasks.
% We evaluate MemBART on three dialogue datasets and two language modeling datasets.

\subsection{Datasets Details}

\begin{table}[H]
    \centering
    \resizebox{0.48\textwidth}{!}{
        \begin{tabular}{l|ccc}
        \toprule
        \textbf{Datasets} & \textbf{\#Turns} & \textbf{Avg. Len} & \textbf{Max Len} \\
        \midrule
        PersonaChat & 14.66 & 244 & 715 \\
        Persuasion & 20.58 & 456 & 1,437 \\
        Multi-Session Chat & 60.52 & 1,823 & 2,705 \\
        \midrule
        Arxiv & - & 13,409 & 156,605 \\
        PG19  & - & 105,830 & 1,181,156 \\
        \bottomrule
        \end{tabular}
    }
    \caption{Dialogue and long document datasets statistics.}
    \label{tab:datasets}
\end{table}

\begin{table*}[h]
    \centering
    \resizebox{1.0\textwidth}{!}{
        \begin{tabular}{lc|c|cccccc}
            \toprule
            \textbf{Base Models} & \textbf{Context} & \textbf{Latency (ms)} $\downarrow$ & \textbf{Total} $\downarrow$ & \textbf{Session 1} $\downarrow$ & \textbf{Session 2} $\downarrow$ & \textbf{Session 3} $\downarrow$ & \textbf{Session 4} $\downarrow$ & \textbf{Session 5} $\downarrow$ \\
            \midrule
            BART base & 128 & 16.41 & 13.05&	10.99&	12.52&	13.18&	13.65&	14.02 \\
            BART base & 256 & 22.12 & 12.83&	10.94&	12.29&	12.97&	13.37&	13.78 \\
            BART base & 512 & 36.80 & 12.68&	10.92&	12.14&	12.77&	13.19&	13.61\\
            BART base & 1,024 & 64.65 & 12.53&	10.81&	11.93&	12.50&	13.10&	13.55 \\
            % Longformer base & 256 & & 12.87&	10.78&	12.36&	13.02&	13.45&	13.88 \\
            % Longformer base & 512 & & 12.69&	10.77&	12.19&	12.79&	13.22&	13.668\\
            % Longformer base & 1,024 & & 12.55&	10.74&	12.12&	12.59&	13.02&	13.48\\
            LED base & 2,048 & 227.75 &12.52&	10.76&	12.13&	12.59&	12.93&	13.42 \\
            % \midrule
            % MemBART base (64) & 128 & & 12.42&	10.72&	11.95&	12.52&	12.93&	13.23 \\
            % MemBART base (64) & 256 & & 12.34&	10.66&	11.86&	12.46&	12.84&	13.16 \\
            % MemBART base (64) & 512 & & 12.23&	10.66&	11.78&	12.32&	12.66&	13.02 \\
            \midrule
            MemBART base (128) & 128 & 20.42 & 12.41&	10.72&	11.95&	12.52&	12.88&	13.23 \\
            MemBART base (128) & 256 & 32.09 & \underline{12.25} &	\textbf{10.62} & \underline{11.76} &	\underline{12.37} &	\underline{12.71} &	\underline{13.06} \\
            MemBART base (128) & 512 & 66.70 & \textbf{12.15} &	\underline{10.63}&	\textbf{11.67}&	\textbf{12.23}&	\textbf{12.57} &	\textbf{12.97} \\
            % \midrule
            % MemBART base (256) & 128 & & 12.38&	10.67&	11.90&	12.51&	12.86 & 13.20\\
            % MemBART base (256) & 256 & & 12.25&	10.59&	11.76&	12.38&	12.74& 13.07 \\
            % MemBART base (256) & 512 & & 12.09&	10.57&	11.62&	12.18&	12.53& 12.90 \\
            \bottomrule
            \toprule
            \textbf{Large Models} & \textbf{Context} & \textbf{Latency (ms)} & \textbf{Total} & \textbf{Session 1} & \textbf{Session 2} & \textbf{Session 3} & \textbf{Session 4} & \textbf{Session 5} \\
            \midrule
            % GPT2-12 & 128 & & 14.36&	12.91&	13.80&	14.43&	14.79&	15.22 \\
            % GPT2-12 & 256 & & 14.13&	12.80&	13.57&	14.21&	14.53&	14.93 \\
            GPT2-12 & 512 & 65.77 & 13.99&	12.81&	13.45&	14.03&	14.33&	14.78 \\
            GPT2-12 & 1,024 & 149.05 & 13.56&	12.82&	13.48&	13.84&	13.53&	13.82 \\
            \midrule
            % GPT2-24 & 128 & & 12.03&	11.17&	11.52&	12.07&	12.30&	12.62 \\
            % GPT2-24 & 256 & & 11.78&	11.02&	11.28&	11.82&	12.04&	12.36 \\
            GPT2-24 & 512 & 172.43 & 11.65&	11.07&	11.14&	11.66&	11.86&	12.20 \\
            GPT2-24 & 1,024 & 395.84 & 11.56& 11.03&	11.12&	11.52&	11.75&	12.11\\
            \midrule
            BART large & 128 & 45.37 & 10.61&	9.50& 10.13&	10.68&	10.94&	11.29 \\
            BART large & 256 & 63.79 & 10.37&	9.38&	9.86&	10.44&	10.67&	11.02 \\
            BART large & 512 & 103.20 & 10.23&	9.44&	9.71&	10.26&	10.52&	10.85 \\
            BART large & 1,024 & 190.79 & 10.10&	9.41&	9.64&	10.06&	10.36&	10.68\\
            % \midrule
            % Longformer large & 256 & & 10.43&	9.34&	9.95&	10.52&	10.75&	11.11 \\
            % Longformer large & 512 & & 10.28&	9.37&	9.77&	10.32&	10.57&	10.92 \\
            % Longformer large & 1,024 & & 10.13&	9.42&	9.66&	10.11&	10.38&	10.72 \\
            LED large & 2,048 & 655.19  & \underline{10.05} &	9.43&	\underline{9.60} &	\underline{10.04} &	\underline{10.27} &	\underline{10.60} \\
            \midrule
            MemBART large (128) & 128 & 59.51 & 10.17&	9.22&	9.61&	10.24&	10.47&	10.85 \\
            MemBART large (128) & 256 & 102.42 & 10.09&	\textbf{9.20}&	9.65&	10.09&	10.38&	10.72\\
            MemBART large (128) & 512 & 197.79 & \textbf{9.99} &	\underline{9.22} &	\textbf{9.51} &	\textbf{10.03} &	\textbf{10.23} &	\textbf{10.58} \\
            \bottomrule
        \end{tabular}
    }
    \caption{MSC perplexity results on the test set. MemBART is able to achieve lower latency while having better performance. Session 4 and session 5 only exist during inference.
    * MemBART (128) means the memory size is 128.
    More details are in Appendix~\ref{appx:msc}}
    \label{tab:main_msc}
\end{table*}

% Dialogues
We experimented on three different dialogue datasets: PersonaChat \cite{DBLP:conf/acl/KielaWZDUS18}, PersuasionForGood \cite{wang-etal-2019-persuasion}, and Multi-Session Chat (MSC) \cite{xu-etal-2022-beyond}.
Especially, Multi-Session Chat addresses the problem of lacking long-context dialogue datasets in the current community.
It is the largest human-human dataset for long conversations with five sessions and average 60 turns of utterances.
% Language modeling
To further test the model's capability, we also evaluate our model on two language modeling tasks: Arxiv and PG19 \cite{DBLP:conf/iclr/RaePJHL20}.
Due to computational constraints, we select the $2,809$ CS AI Arxiv papers, and a subset of 200 books from PG19 for evaluation. We split $10\%$ of the data for testing.
The statistics of all the datasets are shown in Table~\ref{tab:datasets}.

% Evaluation metrics
We compare MemBART with GPT2, BART, and Longformer, as they are all pre-trained language models.  
We use beam search with a beam size of 4 for generation.
For evaluation metrics, we report perplexity and the word overlap F1 for PersonaChat dataset.
For other datasets, we only report perplexity due to the response diversity.
Perplexity reflects the likelihood of the ground truth and it is shown to be highly correlated with other conversation quality metrics.  

% Baselines
% We compare MemBART with GPT2, BART, Longformer.
% We treat GPT2-12 layers as large model based on its computation cost, and this is also discussed in T5 \cite{DBLP:journals/jmlr/RaffelSRLNMZLL20}.

\begin{table}[h]
    \centering
    \resizebox{0.485\textwidth}{!}{
        \begin{tabular}{l|cccc}
        \toprule
\multirow{2}{*}{\textbf{Models}}  & \multicolumn{4}{c}{\textbf{Context Length}} \\ 
& \textbf{128} & \textbf{256} & \textbf{512} & \textbf{1024} \\
\midrule
BART base & 10.93 & 10.90 & 10.80 & 10.78\\
MemBART base (64) & 10.69 & 10.66 & 10.66 & - \\
\quad w/o history & 10.86 & 10.79 & 10.75 & - \\
MemBART base (128) & 10.65 & 10.57 & 10.56 & - \\
MemBART base (256) & \textbf{10.59} & \textbf{10.56} & \textbf{10.54} & - \\
\bottomrule
\toprule
GPT2-12 & 10.51 & 10.38 & 10.33 & 10.31 \\
GPT2-24 & 9.37 & 9.20 & 9.14 & 9.11 \\
BART large & 9.54 & 9.40 & 9.24 & 9.27  \\
MemBART large (128) & \textbf{9.34} & \textbf{9.18} & \textbf{9.12} & - \\
        \bottomrule
        \end{tabular}
    }
    \caption{Perplexity $\downarrow$ results for Persuasion dataset. * MemBART (64) means the memory size is 64.}
    \label{tab:persuasion}
\end{table}

\subsection{Dialogue Datasets Results}

Table~\ref{tab:personachat},\ref{tab:persuasion},\ref{tab:main_msc} show the results for PersonaChat, PersuasionForGood, and MSC, respectively.
We list several main observations below.

\textbf{The memory module memorizes the history information, and the pre-training is necessary.}
In Table~\ref{tab:personachat}, we show that by resetting the memory states (w/o history), MemBART performs similarly to BART base. 
Also, without pre-training, the memory module does not initially learn to memorize the history information.

\textbf{MemBART can be much faster with a small input context size while having better performance.} 
% We can also observe less performance loss due to the truncation of the context size.
In PersonaChat, MemBART with 64 memory size and 64 context length can be on par with the performance of BART with 512 context length.
The same pattern holds for PersuasionForGood (Persuasion) and Multi-Session Chat(MSC) dataset.
Especially in MSC, MemBART base can achieve similar perplexity (12.41) compared to LED base with context length 2,048, but \textit{\textbf{11.15 times faster}}.
MemBART large achieves similar perplexity (10.09) compared to LED large with context length 2,048, while \textit{\textbf{6.40 times faster}}.

\textbf{Encoder-decoder models utilize history information better than decoder-only models.}
For PersonaChat and MSC, BART base and MemBART large outperforms GPT2-12 and GPT2-24 respectively.
The exception is in Persuasion, where the conversations contain more single-turn utterances.
This observation suggests that encoder-decoder models  utilize history information better, and it is probably because of the bidirectional context.

\textbf{MemBART's performance improves as the context size increases.}
BART and GPT2's performance improves when context size increases. 
The results show that increasing the context size for MemBART can also improve its performance, although only by a small margin.
We suspect that using a larger context size can help the model to enhance the memorization of history information and alleviate situations where some information is not kept in the memory.

\textbf{Increasing memory size improves MemBART performance.}
For MemBART models, the history information is stored inside memory.
Thus, we want to study how the performance scales with the memory size.
We evaluated memory size 64, 128, and 256.
We observe that when increasing the size of memory from 64 to 128, there is a large improvement, but from 128 to 256, the improvement is marginal.

\subsection{Language Modeling Datasets Results}

\begin{table}[h]
    \centering
    \resizebox{0.46\textwidth}{!}{
        \begin{tabular}{lc|cc}
        \toprule
\textbf{Models} & \textbf{Context} & \textbf{Arxiv} & \textbf{PG19} \\
\midrule
BART base & 512 & 15.40 & 33.70  \\
BART base & 1,024 & 15.09 & 31.20 \\
LED base & 2,048 & \textbf{13.97} & \underline{30.08}  \\
MemBART base (128) & 512 & \underline{14.34} & \textbf{29.81}  \\
\bottomrule
\toprule
GPT2-12 & 512 & 17.53 & 32.20  \\
GPT2-12 & 1,024 & 15.35 & 28.31 \\
\midrule
GPT2-24 & 512 &15.34 & 22.33 \\
GPT2-24 & 1,024 & 13.84 & \textbf{20.86} \\
\midrule
BART large & 512 & 12.92 & 24.08 \\
BART large & 1,024 & 12.31 &  23.07 \\
LED large & 2,048 & \textbf{11.82} & 23.04 \\
MemBART large (128) & 512 & \underline{12.24} & \underline{22.26} \\
        \bottomrule
        \end{tabular}
    }
    \caption{Language Modeling perplexity scores on Arxiv and PG19 datasets. Lower is better.}
    \label{tab:language_model}
\end{table}

% Motivation
We have also evaluated on two language modeling tasks Arxiv and PG19 to better understand the model's effectiveness. 
% Details of are in Appendix~\ref{appx:pre-training}.
% Details
Due to the computational constraint, we use subsets of the two datasets for evaluation.
We show the results in Table~\ref{tab:language_model}.

% Observation
MemBART performs slightly worse than LED large with 2048 context on Arxiv, but better on PG19.
We suspect that it is because Arxiv papers are very structured and use terminologies across the paper, but PG19 books have less long-term dependency.
The similar performance pattern can also be observed between BART and GPT, which suggests that encoder models are better at using long-term information, and decoder models are better at short-term information.

% On this dataset, MemBART is only able to outperform BART models with 1,024 tokens, but is slightly worse than Longformer with 2,048 context tokens.
% For PG19 books, the average sequence length is much longer than Arxiv papers. However, it 
% MemBART models are able to outperform the Longformer with 2,048 tokens.

\subsection{Ablation Studies}

% % Per-training and history effect
% We analyze the memory's effectiveness under situations.
% In Table~\ref{tab:personachat}, we have shown that without pre-training the memory module, the model performs similarly to BART, which highlights the importance of pre-training.
% Similar pattern happens when the memory is reset at every timestep (w/o history), showing the memory's effectiveness.

% \begin{figure}[!htb]
%     \centering
%     \includegraphics[width=0.4\columnwidth]{images/memory_size.png}
%     \caption{Effect of MemBART's memory size.}
%     \label{fig:ablation_memory_size}
% \end{figure}
% \begin{figure}[!htb]
%     \centering
%     \includegraphics[width=0.4\columnwidth]{images/time_horizon.png}
%     \caption{Effect of back-propagation time horizon}.
%     \label{fig:ablation_time_horizon}
% \end{figure}

% \begin{figure}[!htb]
%   \centering
%   \begin{minipage}[b]{0.4\textwidth}
%     \includegraphics[width=0.4\columnwidth]{images/memory_size.png}
%     % \caption{Effect of MemBART's memory size.}
%     \label{fig:ablation_memory_size}
%   \end{minipage}
%   \hfill
%   \begin{minipage}[b]{0.4\textwidth}
%     \includegraphics[width=0.4\columnwidth]{images/time_horizon.png}
%     % \caption{Effect of back-propagation time horizon}.
%     \label{fig:ablation_time_horizon}
%   \end{minipage}
% \end{figure}

\begin{figure}[!htb]
\centering
\begin{subfigure}{.24\textwidth}
  \centering
  \includegraphics[width=0.95\linewidth]{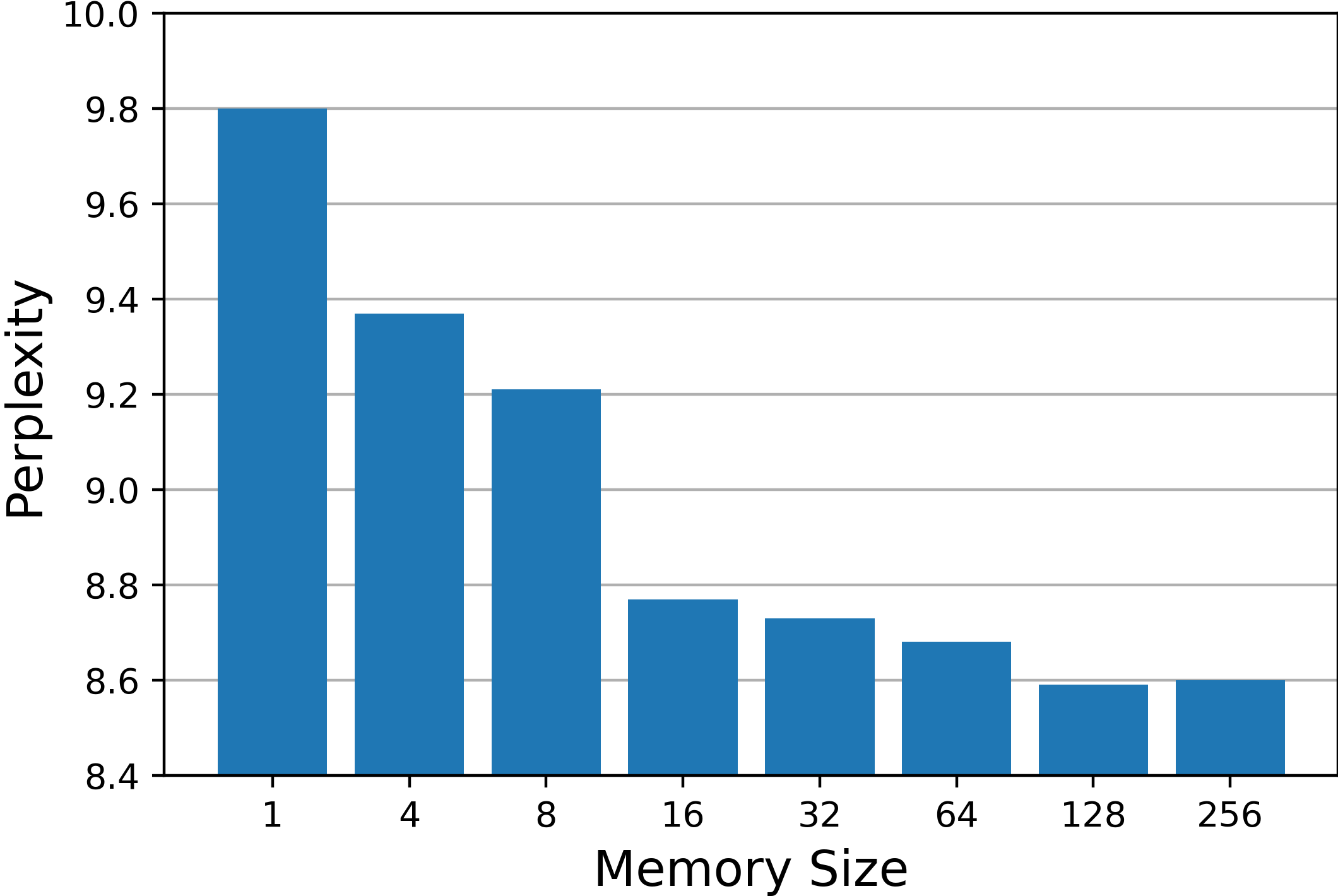}
%   \caption{A subfigure}
%   \label{fig:ablation_memory_size}
\end{subfigure}%
\begin{subfigure}{.24\textwidth}
  \centering
  \includegraphics[width=0.95\linewidth]{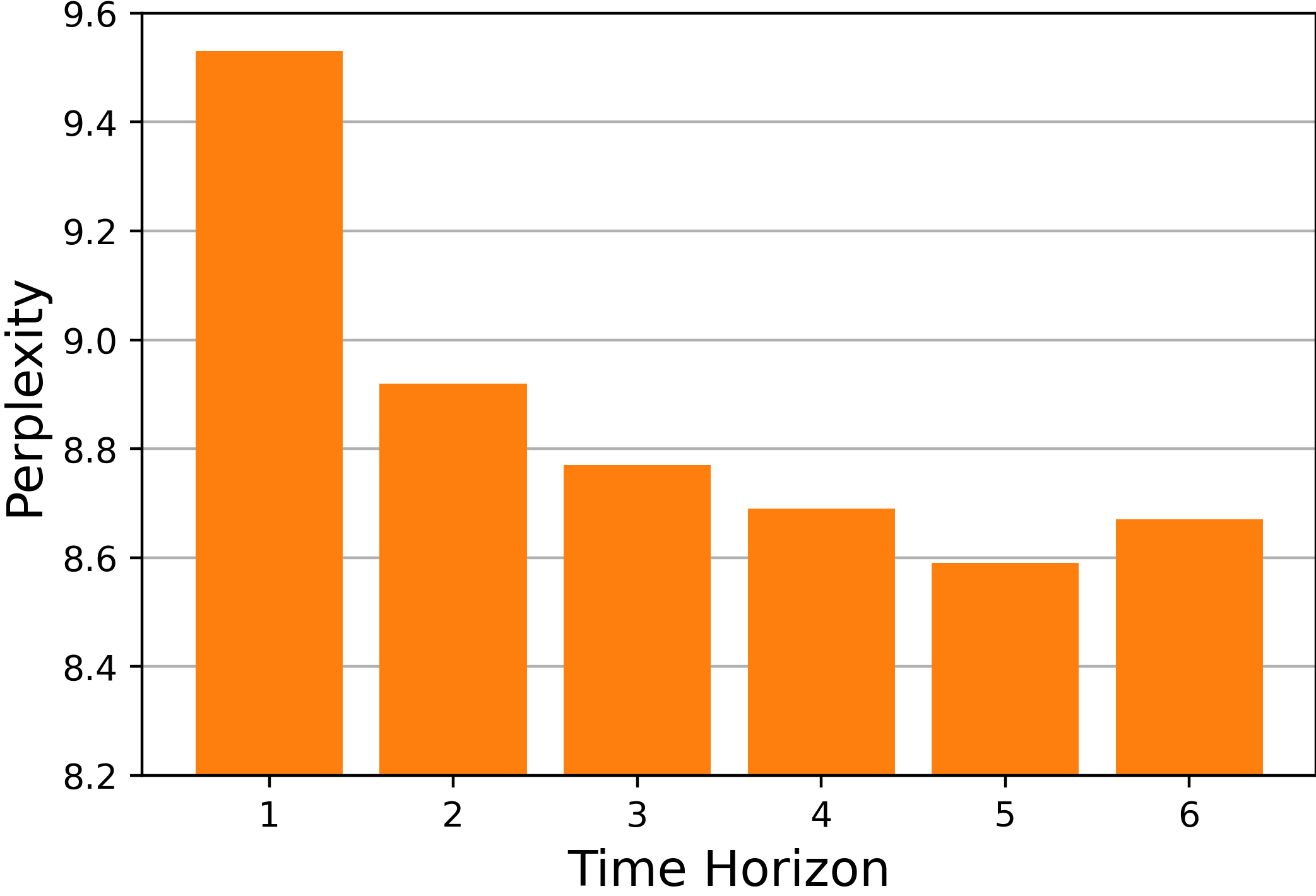}
%   \caption{A subfigure}
%   \label{fig:ablation_time_horizon}
\end{subfigure}
\caption{Effects of changing memory size (left) and time horizon (right).}
\label{fig:ablation}
\end{figure}

We also evaluate the effect of varying memory sizes and back-propagation time horizons on PersonaChat dataset with a context length of 64.
When varying the memory size, we set the time horizon to 5.
In Figure~\ref{fig:ablation}, increasing the memory size has a significant improvement for perplexity until it reaches 128.
When varying the time horizon, memory size is set to 128.
In the right figure, the time horizon being 1 (gradients cannot flow through memory) achieved performance better than BART, suggesting that the memory after pre-training can capture history information.
Increasing the time horizon to 2 can significantly improve the performance.

% Our method has increased number of parameters.
% It slightly improves the result, but overall compared to MemBART with memory, the performance is much worse.

\section{Conclusion}

% What we do
In conclusion, we have introduced a new stateful memory-augmented Transformer encoder-decoder model that can preserve long dialogue history while being compatible with pre-trained encoder-decoder models.
% What we do in detail
By incorporating a separate memory module with dual attention stream and residual gating mechanism, our model effectively interchanges information between the memory states and the pre-trained transformer.
% results
The experimental results have demonstrated the superiority of our method in terms of efficiency and performance, when comparing with other pre-trained models such as BART, GPT, and Longformer (LED).

% Future work
For future work, we plan to broaden the compatibility of our approach with other pre-trained models, and evaluate its performance on other tasks such as task-oriented dialogue systems, text summarization, and long-document classification. Also, we will investigate more advanced memory representations to further optimize the efficiency of the current model.

\newpage

\bibliography{anthology,custom}

\newpage
\appendix

\section{Different Model Variants}
\label{appx:text_recall_pretrain}

We evaluate different model variants to select the model with best memory effectiveness.
% Talk about the task
We choose the text recall task for evaluation.
The task is constructed as recalling previous text segment.
Suppose we have an a document split into text segments $x_0, x_1, \ldots, x_t$.
The encoder receives an input $x_t$ at timestep $t$.
The decoder needs to predict $x_{t-1}$.
In this way, memory has to compress the previous information into the memory.

\textbf{Memformer} The first model is directly applying Memformer by adding the memory cross-attention layers to BART.
The cross-attention layer is between the attention layer and the MLP layer.
Below is the simplified formulation without showing the normalization:
\begin{align*}
    H^l &= H^l + \text{Attn}(H^l) \\
    H^l &= H^l + \text{CrossAttn}(H^l, M_t) \\
    H^l &= H^l + \text{MLP}(H^l)
\end{align*}

\textbf{Memformer + ReZero} uses ReZero \cite{DBLP:conf/uai/BachlechnerMMCM21} by adding a zero-initialized trainable weight $\alpha$ when adding the memory cross-attention layer, and therefore the model's output distribution will get updated smoothly.
\begin{align*}
    H^l &= H^l + \text{Attn}(H^l) \\
    H^l &= H^l + \alpha \, \text{CrossAttn}(H^l, M_t) \\
    H^l &= H^l + \text{MLP}(H^l)
\end{align*}

\textbf{MemBART + Shared weights} A direct variant of our approach is sharing the weights between the memory module and the pre-trained Transformer.
This is similar to append trainable prompting embeddings to the input sequence. 

\textbf{MemBART} is our proposed approach.
The main difference from Memformer is the memory module, where the memory reading and writing are handled with a separate Transformer.
The information flow between the memory module and the pre-trained Transformer is achieved by the dual attention flow to minimally influence the original model distribution.

\begin{table}[!htb]
    \centering
    \resizebox{0.38\textwidth}{!}{
        \begin{tabular}{l|c}
            \toprule
            \textbf{Hyperparams}  & \quad \textbf{All models} \quad \\
            \midrule
            Encoder Layers & 6  \\
            Decoder Layers & 6  \\
            Hidden size & 768  \\
            Attention heads & 12 \\
            Memory size & 32 \\
            \midrule
            Context length & 512 \\
            Batch size       & 8  \\
            Warm-up steps    & 1k   \\
            Learning rate      & 3e-5  \\
            Time horizon & 2  \\
            Dropout           & 0.0    \\
            Weight decay      & 0.01  \\
            Maximum Update steps    & 100k   \\
            \bottomrule
        \end{tabular}
    }
    \caption{Hyper-parameters for the text recall task.}
    \label{tab:text_recall_hyperparameter}
\end{table}

The detailed training hyper-parameters are shown in the Table~\ref{tab:text_recall_hyperparameter}. 
The back-propagation time horizon is set to 2 because it is sufficient for this task. The training takes approximately less than 12 hours to finish on one A6000 GPU.

\section{Sequence Denoising Pre-training Details}
\label{appx:pre-training}

As mentioned, we use the same training objective as BART.
Also, the pre-training corpus is selected to similar to BART.
Since our model is highly based on BART, we use the same tokenization as BART.
We filter out documents that are shorter than $512$ tokens.
% Hyper-parameters
Each document is split into segments with a window size of $512$ and an overlap of $128$ tokens.

\begin{table}[!htb]
    \centering
    \resizebox{0.48\textwidth}{!}{
        \begin{tabular}{l|cc}
            \toprule
             \textbf{Hyperparams} & \textbf{MemBART-base} & \textbf{MemBART-large} \\
            \midrule
            Encoder Layers & 6 & 12 \\
            Decoder Layers & 6 & 12 \\
            Hidden size & 768 & 1024 \\
            Attention heads & 12 & 16 \\
            \midrule
            Context length & 512 & 512 \\
            Stride & 128 & 128 \\
            mask ratio & 0.3 & 0.3 \\
            permutation ratio & 0.0 & 0.0 \\
            replace length & 1 & 1 \\
            \midrule
            Batch size       & 32 & 32 \\
            Warm-up steps    & 5k & 5k  \\
            Learning rate      & 3e-5 & 1e-5 \\
            Time horizon & 6 & 6 \\
            Dropout           & 0.0 & 0.0   \\
            Weight decay      & 0.01 & 0.01 \\
            Update steps    & 100k & 100k   \\
            \bottomrule
        \end{tabular}
    }
    \caption{Hyper-parameters for training MemBART-base and MemBART-large.}
    \label{tab:hyper_denoising}
\end{table}

We pre-train our models with the hyper-parameters shown in Table~\ref{tab:hyper_denoising}.
The pre-training for MemBART-base takes about 4 day on four A6000 GPUs.
The pre-training for MemBART-large takes about 8 days on four A6000 GPUs.

\subsection{Batch Processing and Dispatch}
\label{appx:batch-processing}

% Data dispatcher
\begin{figure}[h]
    \centering
    \includegraphics[width=0.48\textwidth]{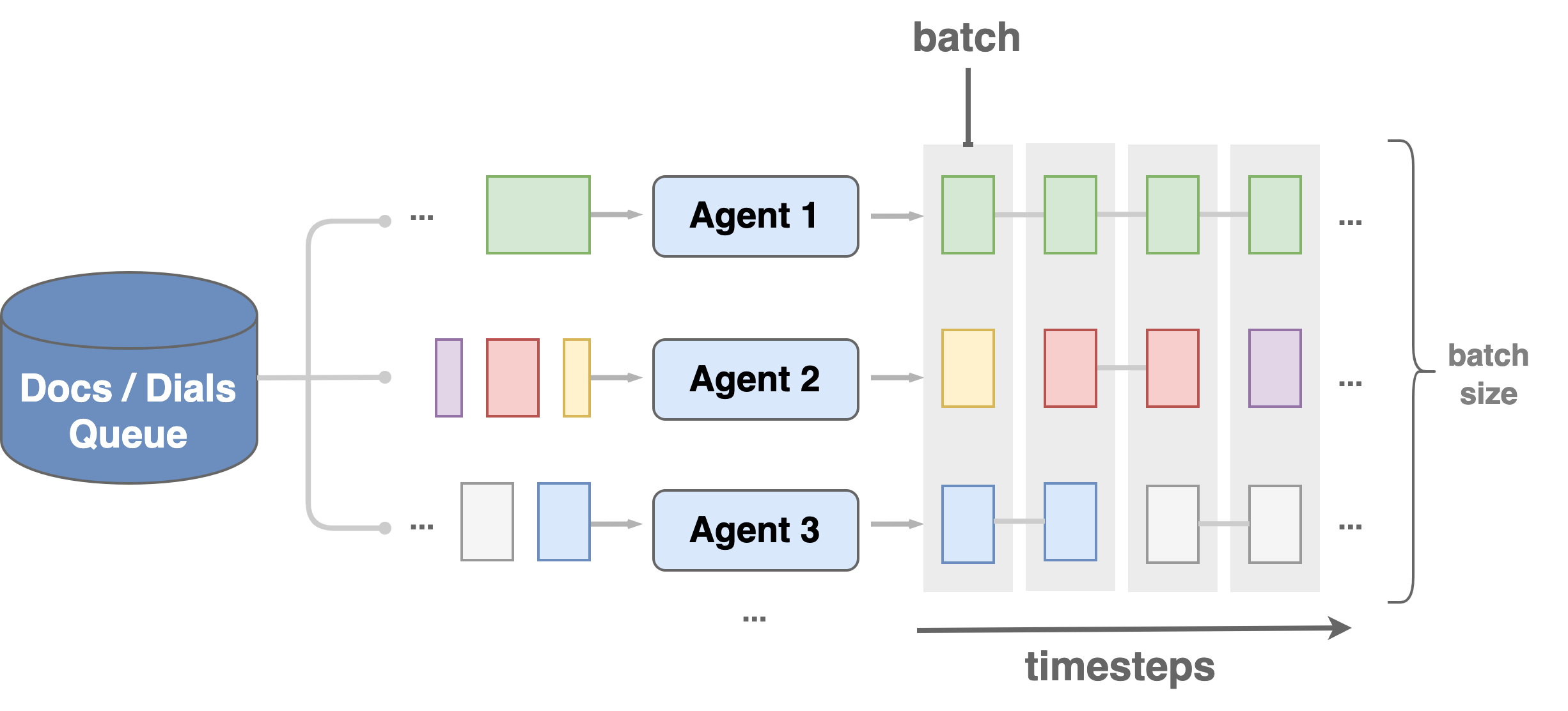}
    \caption{The illustration of how documents or dialogues are processed and batched.}
    \label{fig:data_dispatcher}
\end{figure}

% Motivation
As batches are temporal-dependent in our paradigm, we implement a batch dispatcher to efficiently process the documents and dialogues as shown in Figure~\ref{fig:data_dispatcher}.
% How we do it
In this paradigm, a number of the agents whose size is equal to the batch size share the same data queue to fetch documents. 
When finished processing a document, the agent pops a new document from the shared queue, and it splits the document into text segments or utterances to output one context input at each timestep.
The agent also handles the reset signal and token padding when documents have varied lengths.
All the agents are synchronized, and the batch is collected at each timestep.
% Benefit
This paradigm simplifies the preservation of the temporal order in batches and the alignment between varied-length documents or dialogues.
We use this batch dispatcher across all our experiments.

\section{Multi-Session Chat Full Experiments}
\label{appx:msc}

We have shown the full experiments on multi-session chat under different settings.
Latency is measured with dummy inputs. 
We report the average of 10 runs and the corresponding variance.
We select the best models based on the validation set and then evaluate them on the test set.
The validation results are shown in Table~\ref{tab:msc_valid}.
The test results are shown in Table~\ref{tab:msc_test}.

One observation is that Longformer would pad the sequence to the multiples of $1,024$ due to the sparse attention mechanism.
This behavior results in very slow performance when the context size is small.

Another observation is that for later sessions, especially Session 4 and 5, history information matters.
For Session 5, BART base gets $4.5\%$ performance loss when the context size is truncated to $128$.
BART large gets $6.5\%$ performance loss due to truncation.
In contrast, as MemBART has memory, the performance difference is smaller when using different context sizes.

% \section{Qualitative Analysis}

\section{The Number of Parameters}

\begin{table}[h]
    \centering
    \resizebox{0.35\textwidth}{!}{
        \begin{tabular}{l|c}
            \toprule
             \textbf{Models} & \textbf{\#Parameters}  \\
            \midrule
            BART base & 139M \\
            MemBART base & 183M \\
            \midrule
            BART large & 406M \\
            MemBART large &  558M \\
            \bottomrule
        \end{tabular}
    }
    \caption{The number of parameters for BART and MemBART.}
    \label{tab:parameters}
\end{table}

We show the number of parameters of BART and MemBART in Table~\ref{tab:parameters}. Since MemBART incorporates additional memory module. It is slightly larger than its counterpart BART model.
But as a trade-off, MemBART is much faster than BART.

\section{GPU Memory Efficient Training}

% % Memory replay back-propagation
% To train a recurrent neural network normally requires huge memory space by unrolling the computation graph.
% We apply the same algorithm in Memformer with truncated back-propagation through time (TBPTT) and 
% use a variant of gradient checkpointing called memory replay back-propagation (MRBP) to achieve nearly constant memory consumption during training.

Memformer proposed a variant of gradient checkpointing 
% (Algo.~\ref{algo:mrbp}) 
to efficiently train this type of stateful models. 
The GPU memory consumption scales linearly with the back-propagation time horizon because it requires unrolling the computation graph as equal to the number of timesteps.

We applied this efficient training algorithm for the MemBART large model model with time horizon 6.
Without efficient back-propagation method, it would consume a large amount of GPU memory, which makes the training infeasible.

% \SetKwComment{Comment}{$\triangleright$\ }{}

% \LinesNumbered
% \begin{algorithm}[H]
%     \caption{BP through Memory Replay}
%     \label{algo:mrbp}
%     \KwIn{rollout=[$x_t, x_{t+1}, \ldots, x_T$]: \texttt{a list containing previous inputs} \newline
%     memories=[$M_t, M_{t+1}, \ldots, M_T$]: \texttt{memory from the previous}
%     }
%     \Comment{Initialize a list for back-propagation} 
%     replay = list([$M_t$]) \\
%     % \Comment{previous memory} 
%     % replayBuffer.append() \\ 
%     \Comment{Forward pass \& no gradient} 
%     \For {$t=t, t+1, \ldots, T - 1$}
%     {
%         $M_{t+1}$, $\_$ = Model($x_t$, $M_t$) \\
%         replay.append($M_{t+1}$)
%     }
%     \Comment{Backward pass with gradient}
%     $\nabla M_{t+1} = 0$ \\
%     \For {$t=T, T-1, \ldots, t+1, t$}
%     {   \Comment{Recompute}
%         $M_{t+1}$, $O_{t}$ = Model($x_t$, $M_t$, $r_t$) \\
%         $loss$ = $L(O_{t})$ \\
%         $loss$.backward() \\
%         $M_{t+1}$.backward($\nabla M_{t+1}$) \\
%         $\nabla M_{t+1}$ = $\nabla M_{t}$
%     }
%     \Comment{Update the memories}
%     memories = Buffer \\
%     memories.pop()
% \end{algorithm}

\begin{table*}[h]
    \centering
    \resizebox{0.99\textwidth}{!}{
        \begin{tabular}{lc|ccccccc}
            \toprule
            \textbf{Base Models} & \textbf{Context} & \textbf{Latency} & \textbf{Total} & \textbf{Session 1} & \textbf{Session 2} & \textbf{Session 3} & \textbf{Session 4} & \textbf{Session 5} \\
            \midrule
            BART base & 128 & $16.41_{\pm 0.73}$ & 12.72&	10.84&	13.19&	13.15&	13.17&	12.77 \\
            BART base & 256 & $22.12_{\pm 0.89}$ & 12.50&	10.77&	12.85&	12.89&	12.96&	12.58 \\
            BART base & 512 &  $36.80_{\pm 1.17}$ & 12.33&	10.71&	12.61&	12.67&	12.81&	12.43 \\
            BART base & 1,024 & $64.65_{\pm 0.72}$ & 12.22&	10.69&	12.46&	12.38&	12.77&	12.38 \\
            \midrule
            Longformer base & 256 & $110.07_{\pm 0.28}$ & 12.55&	10.78&	12.92&	12.93&	13.07&	12.57 \\
            Longformer base & 512 & $113.73_{\pm 3.16}$ & 12.35&	10.73&	12.64&	12.66&	12.87&	12.40 \\
            Longformer base & 1,024 & $115.96_{\pm 0.25}$ & 12.20&	10.67&	12.55&	12.46&	12.65&	12.26 \\
            Longformer base & 2,048 & $227.75_{\pm 0.13}$ & 12.16& 10.69&	12.54&	12.46&	12.58&	12.15\\
            \midrule
            MemBART base (64) & 128 & $17.23_{\pm 1.19}$ & 12.17& 10.6&	12.60&	12.54&	12.55&	12.14 \\
            MemBART base (64) & 256 & $29.39_{\pm 0.73}$ & 12.06&	10.59&	12.40&	12.36&	12.47&	12.09 \\
            MemBART base (64) & 512 & $59.73_{\pm 0.66}$ & 11.95&	10.57&	12.28&	12.22&	12.33&	11.98 \\
            \midrule
            MemBART base (128) & 128 & $20.42_{\pm 1.47}$ & 12.12& 10.6&	12.50&	12.45&	12.51&	12.14 \\
            MemBART base (128) & 256 & $32.09_{\pm 0.18}$ & 11.96&	10.49&	12.29&	12.28&	12.37&	11.97 \\
            MemBART base (128) & 512 & $66.70_{\pm 1.83}$  & 11.86&	10.50&	12.15&	12.14&	12.27&	11.89 \\
            \midrule
            MemBART base (256) & 128 & $26.56_{\pm 0.57}$  & 12.11&	10.58&	12.51&	12.43&	12.47&	12.13 \\
            MemBART base (256) & 256 & $40.92_{\pm 0.63}$ & 12.00&	10.50&	12.35&	12.34&	12.40&	12.01 \\
            MemBART base (256) & 512 & $75.54_{\pm 0.14}$ & 11.83&	10.47&	12.11&	12.10&	12.24&	11.86 \\
            \toprule
            \textbf{Large Models} & \textbf{Context} & \textbf{Latency} & \textbf{Total} & \textbf{Session 1} & \textbf{Session 2} & \textbf{Session 3} & \textbf{Session 4} & \textbf{Session 5} \\
            \midrule
            GPT2-12 & 128 & $16.24_{\pm 1.13}$ & 14.17 &  12.87 & 14.57 & 14.5 & 14.51 & 14.03 \\
            GPT2-12 & 256 & $30.80_{\pm 0.48}$ & 13.91 &  12.70 & 14.20 &  14.23& 14.25 & 13.81 \\
            GPT2-12 & 512 & $65.77_{\pm 0.74}$ & 13.76 &  12.68 & 14.03 & 14.02 & 14.11 & 13.67 \\
            GPT2-12 & 1,024 & $149.05_{\pm 0.38}$ & 13.33 & 12.66 & 14.04 & 13.82 & 13.26 & 12.71 \\
            \midrule
            GPT2-24 & 128 & $42.39_{\pm 2.50}$ & 11.91 & 11.15 &	12.17 &	12.10 &	12.10 &	11.83 \\
            GPT2-24 & 256 & $81.80_{\pm 0.18}$ & 11.66 &	10.98&	11.83&	11.83&	11.86&	11.62\\
            GPT2-24 & 512 & $172.43_{\pm 0.12}$ & 11.52&	10.99&	11.63&	11.64&	11.72&	11.48\\
            GPT2-24 & 1,024 & $395.84_{\pm 0.64}$ & 11.43 & 10.96&	11.59&	11.48&	11.62&	11.37\\
            \midrule
            BART large & 128 & $45.37_{\pm 1.31}$ & 10.42&	9.31&	10.75&	10.61&	10.68&	10.44\\
            BART large & 256 & $63.79_{\pm 0.40}$ & 10.15&	9.17&	10.35&	10.34&	10.40&	10.20  \\
            BART large & 512 & $103.20_{\pm 2.40}$ & 10.00&	9.22&	10.12&	10.12&	10.28&	10.03\\
            BART large & 1,024 & $190.79_{\pm 0.29}$ & 9.87&	9.20&	10.03&	9.91&	10.09&	9.90\\
            \midrule
            Longformer large & 256 & $316.42_{\pm 2.37}$ & 10.25&	9.28&	10.43&	10.41&	10.55&	10.30 \\
            Longformer large & 512 & $322.68_{\pm 1.74}$ & 10.06&	9.24&	10.18&	10.15&	10.38&	10.13 \\
            Longformer large & 1,024 & $334.87_{\pm 5.54}$ & 9.90 &	9.20&	10.06&	9.95&	10.15&	9.92 \\
            Longformer large & 2,048 & $655.19_{\pm 5.25}$ & 9.87 &	9.23&	10.09&	9.90&	10.04&	9.89 \\
            \midrule
            MemBART large (128) & 128 & $59.51_{\pm 0.91}$ & 9.99 &	9.17&	10.19&	10.14&	10.22&	10.02 \\
            MemBART large (128) & 256 & $102.42_{\pm 2.07}$ & 9.92&	9.08&	10.10&	10.06&	10.15&	9.95 \\
            MemBART large (128) & 512 & $197.79_{\pm 4.85}$ & 9.79&	9.08&	9.90&	9.88&	10.03&	9.84 \\
            \bottomrule
        \end{tabular}
    }
    \caption{Multi-Session Chat results on the validation set.}
    \label{tab:msc_valid}
\end{table*}

\begin{table*}[h]
    \centering
    \resizebox{0.99\textwidth}{!}{
        \begin{tabular}{lc|ccccccc}
            \toprule
            \textbf{Base Models} & \textbf{Context} & \textbf{Latency} & \textbf{Total} & \textbf{Session 1} & \textbf{Session 2} & \textbf{Session 3} & \textbf{Session 4} & \textbf{Session 5} \\
            \midrule
            BART base & 128 & $16.41_{\pm 0.73}$ & 13.05&	10.99&	12.52&	13.18&	13.65&	14.02 \\
            BART base & 256 & $22.12_{\pm 0.89}$ & 12.83&	10.94&	12.29&	12.97&	13.37&	13.78 \\
            BART base & 512 & $36.80_{\pm 1.17}$ & 12.68&	10.92&	12.14&	12.77&	13.19&	13.61\\
            BART base & 1,024 & $64.65_{\pm 0.72}$ & 12.53&	10.81&	11.93&	12.50&	13.10&	13.55 \\
            \midrule
            Longformer base & 256 & $110.07_{\pm 0.28}$ & 12.87&	10.78&	12.36&	13.02&	13.45&	13.88 \\
            Longformer base & 512 & $113.73_{\pm 3.16}$ & 12.69&	10.77&	12.19&	12.79&	13.22&	13.67\\
            Longformer base & 1,024 & $115.96_{\pm 0.25}$ & 12.55&	10.74&	12.12&	12.59&	13.02&	13.48\\
            Longformer base & 2,048 & $227.75_{\pm 0.13}$ &12.52&	10.76&	12.13&	12.59&	12.93&	13.42 \\
            \midrule
            MemBART base (64) & 128 & $17.23_{\pm 1.19}$ & 12.42&	10.72&	11.95&	12.52&	12.93&	13.23 \\
            MemBART base (64) & 256 & $29.39_{\pm 0.73}$ & 12.34&	10.66&	11.86&	12.46&	12.84&	13.16 \\
            MemBART base (64) & 512 & $59.73_{\pm 0.66}$ & 12.23&	10.66&	11.78&	12.32&	12.66&	13.02 \\
            \midrule
            MemBART base (128) & 128 & $20.42_{\pm 1.47}$ & 12.41&	10.72&	11.95&	12.52&	12.88&	13.23 \\
            MemBART base (128) & 256 & $32.09_{\pm 0.18}$ & 12.25&	10.62&	11.76&	12.37&	12.71&	13.06 \\
            MemBART base (128) & 512 & $66.70_{\pm 1.83}$ & 12.15&	10.63&	11.67&	12.23&	12.57&	12.97 \\
            \midrule
            MemBART base (256) & 128 & $26.56_{\pm 0.57}$ & 12.38&	10.67&	11.90&	12.51&	12.86 & 13.20\\
            MemBART base (256) & 256 & $40.92_{\pm 0.63}$  & 12.25&	10.59&	11.76&	12.38&	12.74& 13.07 \\
            MemBART base (256) & 512 & $75.54_{\pm 0.14}$ & 12.09&	10.57&	11.62&	12.18&	12.53& 12.90 \\
            \toprule
            \textbf{Large Models} & \textbf{Context} & \textbf{Latency} & \textbf{Total} & \textbf{Session 1} & \textbf{Session 2} & \textbf{Session 3} & \textbf{Session 4} & \textbf{Session 5} \\
            \midrule
            GPT2-12 & 128 & $16.24_{\pm 1.13}$ & 14.36&	12.91&	13.80&	14.43&	14.79&	15.22 \\
            GPT2-12 & 256 & $30.80_{\pm 0.48}$ & 14.13&	12.80&	13.57&	14.21&	14.53&	14.93 \\
            GPT2-12 & 512 & $65.77_{\pm 0.74}$ & 13.99&	12.81&	13.45&	14.03&	14.33&	14.78 \\
            GPT2-12 & 1,024 & $149.05_{\pm 0.38}$ & 13.56&	12.82&	13.48&	13.84&	13.53&	13.82 \\
            \midrule
            GPT2-24 & 128 & $42.39_{\pm 2.50}$ & 12.03&	11.17&	11.52&	12.07&	12.30&	12.62 \\
            GPT2-24 & 256 & $81.80_{\pm 0.18}$ & 11.78&	11.02&	11.28&	11.82&	12.04&	12.36 \\
            GPT2-24 & 512 & $172.43_{\pm 0.12}$ & 11.65&	11.07&	11.14&	11.66&	11.86&	12.20 \\
            GPT2-24 & 1,024 & $395.84_{\pm 0.64}$ & 11.56& 11.03&	11.12&	11.52&	11.75&	12.11\\
            \midrule
            BART large & 128 & $45.37_{\pm 1.31}$ & 10.61&	9.50&	10.13&	10.68&	10.94&	11.29 \\
            BART large & 256 & $63.79_{\pm 0.40}$ & 10.37&	9.38&	9.86&	10.44&	10.67&	11.02 \\
            BART large & 512 & $103.20_{\pm 2.40}$ & 10.23&	9.44&	9.71&	10.26&	10.52&	10.85 \\
            BART large & 1,024 & $190.79_{\pm 0.29}$ & 10.10&	9.41&	9.64&	10.06&	10.36&	10.68\\
            \midrule
            Longformer large & 256 & $316.42_{\pm 2.37}$ & 10.43&	9.34&	9.95&	10.52&	10.75&	11.11 \\
            Longformer large & 512 & $322.68_{\pm 1.74}$ & 10.28&	9.37&	9.77&	10.32&	10.57&	10.92 \\
            Longformer large & 1,024 & $334.87_{\pm 5.54}$ & 10.13&	9.42&	9.66&	10.11&	10.38&	10.72 \\
            Longformer large & 2,048 & $655.19_{\pm 5.25}$ & 10.05&	9.43&	9.60&	10.04&	10.27&	10.60 \\
            \midrule
            MemBART large (128) & 128 & $59.51_{\pm 0.91}$ & 10.17&	9.22&	9.61&	10.24&	10.47&	10.85 \\
            MemBART large (128) & 256 & $102.42_{\pm 2.07}$ & 10.09&	9.20&	9.65&	10.09&	10.38&	10.72\\
            MemBART large (128) & 512 & $197.79_{\pm 4.85}$ & 9.99&	9.22&	9.51&	10.03&	10.23&	10.58 \\
            \bottomrule
        \end{tabular}
    }
    \caption{Multi-Session Chat results on the test set.}
    \label{tab:msc_test}
\end{table*}

\end{document}